\documentclass{article}

\usepackage[preprint, nonatbib]{nips_2018}

\usepackage[utf8]{inputenc}
\usepackage[T1]{fontenc}
\usepackage{hyperref}
\usepackage{url}
\usepackage{booktabs}
\usepackage{amsfonts}
\usepackage{nicefrac}
\usepackage{microtype}
\usepackage{paralist}

\usepackage{algorithm}
\usepackage[noend]{algorithmic}
\usepackage{amsmath}
\usepackage{amssymb}
\usepackage{amsthm}
\usepackage{color}
\usepackage{xcolor}
\usepackage{tcolorbox}
\usepackage{mathrsfs}
\usepackage{dsfont}

\usepackage{graphicx}
\usepackage[export]{adjustbox}
\usepackage{subcaption}
\usepackage{caption}
\usepackage{array}

\usepackage{rotating}
\usepackage{multirow}

\graphicspath{{./figures/}}

\title{Adversarially Robust Training through \\ Structured Gradient Regularization}

\author{
  Kevin Roth \\
   Department of Computer Science \\
   ETH Z\"urich\\
  \texttt{\small kevin.roth@inf.ethz.ch} \\
  \And
   Aurelien Lucchi \\
   Department of Computer Science \\
   ETH Z\"urich\\
  \texttt{\small aurelien.lucchi@inf.ethz.ch} \\
   \And
   Sebastian Nowozin\\
   Microsoft Research \\
   Cambridge, UK \\
   \texttt{sebastian.Nowozin@microsoft.com} \\
  \And
   Thomas Hofmann \\
   Department of Computer Science \\
   ETH Z\"urich\\
  \texttt{\small thomas.hofmann@inf.ethz.ch} \\   
}

\newcommand{\Laplace}{{\mathop{}\!\mathbin\bigtriangleup}}

\newcommand{\E}{{\mathbf E}}

\newcommand{\x}{{\mathbf x}}

\renewcommand{\Re}{{\mathbb R}}

\newcommand{\Eq}[1]{Eq.\,({#1})}

\newcommand{\bxi}{\boldsymbol{\xi}}

\newcommand{\bphi}{\boldsymbol{\phi}}
\newcommand{\bSigma}{{\boldsymbol{\Sigma}}}
\newcommand{\bmu}{{\boldsymbol{\mu}}}

\renewcommand{\P}{{\mathbb P}}

\newcolumntype{S}{>{\centering\arraybackslash} m{.08\linewidth} }
\newcolumntype{T}{>{\centering\arraybackslash} m{.10\linewidth} }

\begin{document}

\maketitle

\begin{abstract}
We propose a novel data-dependent structured gradient regularizer to increase the robustness of neural networks vis-a-vis adversarial perturbations. 
Our regularizer can be derived as a controlled approximation from first principles, leveraging the fundamental link between training with noise and regularization. It adds very little computational overhead during learning and is simple to implement generically in standard deep learning frameworks. Our experiments provide strong evidence that structured gradient regularization can act as an effective first line of defense against attacks based on low-level signal corruption.\footnote{Code will be made available at \url{https://github.com/rothk}} 
\end{abstract}


\section{Introduction}
\label{sec:intro}

Deep neural networks, in particular convolutional neural networks (CNNs), have been used with great success for perceptual tasks such as image classification \cite{simonyan2014very} or speech recognition \cite{hinton2012deep}. However, it has been shown that the accuracy of models obtained by standard training methods can dramatically deteriorate in the face of  so-called adversarial examples \cite{szegedy2013intriguing, goodfellow2014explaining}, i.e. small perturbations in the input signal, typically imperceptible to humans, that are sufficient to induce large changes in the output. This apparent vulnerability  is worrisome as CNNs  start to proliferate in the real-world, including in safety-critical deployments.

Although the theoretical aspects of vulnerability to adversarial perturbations are not yet fully understood, a plethora of methods has been proposed to find adversarial examples. These often transfer or generalize across different architectures and datasets \cite{papernot2016transferability, liu2016delving, tramer2017space}, enabling black-box attacks even for inaccessible models. The most direct and commonly used strategy of  protection is to use adversarial examples during training, see e.g.~\cite{goodfellow2014explaining, miyato2015distributional}. This raises the question, of whether one can generalize across such examples in order to become immune to a wider range of possible adversarial perturbations. Otherwise there appears to be a danger of overfitting to a specific attack, in particular if adversarial examples are sparsely generated from the training set. 

In this paper, we pursue a different route, starting from the hypothesis that the weakness revealed by adversarial examples first  and foremost hints at an overfitting or data scarcity problem. Inspired by the classical work of \cite{bishop1995noise}  and similar in spirit to \cite{miyato2017virtual}, we thus propose regularization as the preferred remedy to increase model robustness. Our regularization approach comes without rigid \textit{a~priori} assumptions on the model structure and filters \cite{bruna2013invariant} as well as strong (and often overly drastic) requirements such as layer-wise Lipschitz bounds \cite{cisse2017parseval} or isotropic smoothness \cite{miyato2017virtual}. Instead, we will rely on the ease of generating adversarial examples to learn an \textit{informative} regularizer, focusing on the correlation structure of the adversarial noise. 
Our structured gradient regularizer consistently outperforms structure-agnostic baselines on long-range correlated perturbations, achieving over $25\%$ higher classification accuracy. 
We thus consider our regularization approach a first line of defense on top of which even more sophisticated defense mechanisms may be built.

In summary, we make the following contributions: 
\begin{itemize}
\item We provide evidence that current adversarial attacks act by perturbing the short-range covariance structure of signals.
\item We propose Structured Gradient Regularization (SGR), a data-dependent regularizer informed by the covariance structure of adversarial perturbations. 
\item We present an efficient SGR implementation with low memory footprint.
\item We introduce a long-range correlated adversarial attack against which SGR-regularized classifiers are shown to be robust.
\end{itemize}


\section{Adversarial Robust Learning}

\subsection{Virtual Examples}
Imagine we had access to a set of transformations $\tau \in \mathcal T$ of inputs that leave outputs invariant. Then we could augment our training data by expanding each real example $(\x_i,y_i)$ into \textit{virtual examples} $(\tau(\x_i),y_i)$. This approach is simple and widely-applicable, e.g.~for exploiting known data invariances \cite{scholkopf1996incorporating}, and can greatly increase the statistical power of learning. A similar stochastic method has been suggested by  \cite{maaten2013learning}: a noisy channel or perturbation $Q$ corrupts $\x \mapsto \tilde \x $ with probability $Q(\tilde \x | \x)$, leading to an altered (more precisely: convolved) joint distribution $P_Q(\tilde \x,y)  = P(y) \int P(\x|y) Q(\tilde \x | \x) d\x$. 

Where could such a corruption model come from? We propose to use a generative mechanism to learn adversarial corruptions from examples. In fact, we will use a simple additive noise model in this paper,
\begin{align}
\label{eq:q}
Q_{}(\tilde \x|\x) = Q_{}(\bxi \!:=\! \tilde \x - \x), \;  
\E[ \bxi \bxi^\top] = \bSigma\,.
\end{align}
Our focus will be on leveraging the correlation structure of the noise, captured by the matrix of raw second moments 
$\hat{\bSigma}$, aggregated as a running exponentially weighted batch estimate computed from
adversarial perturbations $\bxi_i$, obtained by methods such as FGSM~\cite{goodfellow2014explaining}, PGD~\cite{madry2017towards} or DeepFool~\cite{moosavi2016deepfool}.

\subsection{Robust Multiclass Learning Objective} 

Once we have an estimate of $Q$ available, we propose to optimize any standard training criterion (e.g.~a likelihood function) in expectation over a mixing distribution between the uncorrupted and the corrupted data. For the sake of concreteness, let us focus on the multiclass case, where $y \in \{1,\dots,K\}$. Define a probabilistic classifier
\begin{align}
\bphi: \Re^d \to \triangle^K, \quad \triangle^K := \{ \pi \in \Re_+^K: \mathbf 1^\top \pi  =1\}\,.
\end{align}
For a given loss function $\ell$  such as the negative log-loss, $\ell(y,\pi) = -\log \pi_y$, our aim is to minimize
\begin{align}
\label{eq:mix-objective}
\mathcal L(\bphi) \!=\! 
	 (1\!-\!\lambda) \E_{\hat P}\left[ \ell (y,\bphi(\x)) \right] 
	\!+\!  \lambda \E_{\hat P_Q} \left[ \ell (y,\bphi(\x)) \right],
\end{align}
where $\hat P$ denotes the empirical distribution and $\hat P_Q$ the noise corrupted distribution. The hyper-parameter $\lambda$ gives us a degree of freedom to pay more attention to the training data vs.~our (imperfect) corruption model.

\subsection{From Virtual Examples to Regularization} 

Approaches relying on the explicit generation of adversarial examples face three \textit{statistical challenges}: (i) How can enough samples be generated to cover all attacks? (ii) How can we avoid a computational blow-up by adding too many virtual examples? (iii) How can we prevent overfitting to a specific attack? A remedy to these problems is through the use of regularization. The basic idea is simple: instead of sampling virtual examples, one tries to calculate the corresponding integrals in closed form, at least under reasonable approximations. 

The general connection between regularization and training with noise-corrupted examples has been firmly established by \cite{bishop1995noise}. The benefits of regularization over adversarially augmented training in terms of generalization error have been hypothesized in \cite{galloway2018adversarial}. In \cite{maaten2013learning}, properties of the loss function are exploited to achieve efficient regularization techniques for certain types of noise. Here, we follow the approach pursued in \cite{roth2017stabilizing} in using an approximation that is accurate for small noise amplitudes.


\section{Structured Gradient Regularization}
\label{sec:derivations}

\subsection{From Correlated Noise to Structured Gradient Regularization}
\label{sec:CorrelatedNoisetoSGR}
To approximate the expectation with regard to the noisy channel, we generalize a recent approach for gradient regularization of binary classifiers in Generative Adversarial Networks (GANs) \cite{roth2017stabilizing}:  we approximate the expectation over corruptions through a Taylor expansion, which subsequently allows for integrating over perturbations.

For the sake of simplicity of  derivations we assume adversarial perturbations to be \textit{centered}. Let us Taylor expand the log-component functions, making use of the shortcut  $\psi_y := \log \phi_y$,
\begin{align}
\label{eq:taylor}
\psi_y(\cdot + \bxi)  =  \psi_y + \nabla\psi_y^\top \bxi + \tfrac 12  \bxi^\top \Laplace[\psi_y] \bxi + \mathcal O(\| \bxi\| ^3)
\end{align}
For any zero-mean distribution $Q$ with second moments $\bSigma$, 
\begin{align}
\E_Q[\psi_y(\cdot + \bxi)] = \psi_y \! +\! \tfrac 12 \text{Tr}(\Laplace[\psi_y] \bSigma) \!+\! \mathcal{O}(\E[\|\bxi\|^3]).
\end{align}
The Hessian is calculated via the chain rule 
\begin{align}
\Laplace(\psi_y) =  \nabla \left[\frac{\nabla \phi_y}{\phi_y}\right]  =
\frac{\Laplace \phi_y}{\phi_y} - \nabla \psi_y \cdot \nabla \psi_y^\top \, .
\end{align}
We will make a similar argument to \cite{roth2017stabilizing} to show that it is reasonable to neglect the terms involving Hessians  $\Laplace \phi_y$.  Let us therefore consider the Bayes-optimal classifier $\phi_y^*$, for which up to a normalization constant
\begin{align}
\phi_y^*(\x) \propto P(\x|y) P(y) \,,
\end{align}
such that 
\begin{align}
\E_{P} \left[ \frac{\Laplace \phi^*_y(\x)}{\phi^*_y(\x)} \right] \propto \int \sum_y \Laplace \phi^*_y(\x) d\x\,.
\end{align}
Exchanging summation and differentiation, we however have as a consequence of the normalization
\begin{align}
\sum_y \Laplace \phi_y(\x) = \Laplace \bigg[ \sum_{y} \phi_y(\x) \bigg] = \Laplace 1 = 0 \,.
\end{align}
Thus, under the assumption that $\phi \approx \phi^*$ and of small perturbations (such that we can ignore higher order terms in Eq.~\eqref{eq:taylor}), we get 
\begin{align}
\E_{P_Q}[ \psi_y(\x)] = 
& \; \E_{P}\left[ \psi_y(\x) - \tfrac 12 \text{Tr}\left[ \nabla \psi_y \nabla \psi_y^\top \bSigma\right] \right] + \mathcal{O}(\E[\|\bxi\|^3]) \, .
\end{align}

We can turn this into a regularizer by taking the leading terms \cite{roth2017stabilizing}.
Identifying $P = \hat P$ with the empirical distribution, 
we arrive at the following robust learning objective:
\begin{tcolorbox}
\vspace{-.35cm}
\begin{align}
\label{eq:ART_regularizer}
& \hspace{-4pt}\vspace{-1mm}\textnormal{\bf Structured Gradient Regularization (SGR)} \nonumber\\[2mm]
&\hspace{-6pt}\min {\bphi}\,\to \mathcal L^{\rm SGR}(\bphi) =  \E_{\hat P} \left[ -\log \phi_y(\x) \right] 
 + \lambda \Omega_{\bSigma}(\bphi) \\[2mm]
&\hspace{-4pt} \Omega_{\bSigma}(\bphi) := \frac{1}{2} \E_{\hat P} \Big[ \nabla_\x \log \phi_y(\x)^\top \, \bSigma \, \nabla_\x \log \phi_y(\x) \Big]
\nonumber
\end{align}
\vspace{-.5cm}
\end{tcolorbox}

For uncentered adversarial perturbations $\E_Q[\bxi]=\bmu$ it is easy to see that the corresponding regularizer is given by
\begin{align}
& \Omega_{\bmu, \bSigma}(\bphi) = - \E_{\hat P}\bigg[ \nabla \log\phi_y^\top(\x) \, \bmu \bigg] 
+ \Omega_{\bSigma}(\bphi) \, .
\end{align}
In practice, we however observed this correction to be small.

\subsection{Properties}
\label{sec:SGRproperties}
There are a few facts that are important to point out about the derived SGR regularizer. 

(i) As the regularizer provides an efficiently computable approximation for an intractable expectation, it is clearly \textit{data-dependent}. $\Omega_{\bSigma}$ penalizes loss-gradients evaluated at training points. This is different from standard regularizers that penalize some norm of the parameter vector, such as $L_2$-regularization. In the latter case, we would expect the regularizer to have the largest effect in the empty parts of the input space, where it should reduce the variability of the classifier outputs. On the contrary, $\Omega_{\bSigma}$ has its main effect around the data manifold. In this sense, it is complementary to standard regularization methods.

(ii) Moreover, the SGR regularizer is \textit{intrinsic} in the sense that it does not depend on the parameters of the classifier (for a good reason, we have not mentioned any parameterization), 
but instead directly acts on the function realized by the classifier $\bphi$
(the relevant gradients measure the sensitivity of $\bphi$ with regard to the input $\x$ and not with regard to parameters).
Thus, it is parameterization invariant and can be naturally applied  to any function space, whether it is finite-dimensional or not.

(iii) We can gain complementary insights into SGR by explicitly computing it in terms of classifier logits $\varphi_y(\x)$.
As outlined in Sec~\ref{sec:SGRpropertiesAppendix} in the Appendix, 
we obtain the following expression for the structured gradient regularizer
\begin{align}
\hspace{-3pt}\Omega_{\bSigma}(\bphi) = \frac{1}{2} \E_{\hat P} \bigg[ ( \nabla\varphi_y \hspace{-1pt}-\hspace{-1pt} \left\langle  \nabla\varphi  \right\rangle )^\top \bSigma ( \nabla\varphi_y \hspace{-1pt}-\hspace{-1pt} \left\langle  \nabla\varphi  \right\rangle ) \bigg] \,\,\, , \left\langle \nabla \varphi  \right\rangle(\x) := \sum_y \nabla \varphi_y(\x) \phi_y(\x)
\end{align}
We can therefore see that SGR is penalizing large \textit{variations} of the class-conditional logit-gradients $\nabla\varphi_y$ around their data-dependent class average $\left\langle  \nabla\varphi  \right\rangle$. 
For simple one-layer softmax classifiers 
we obtain
$ \Omega_{\bSigma}(\bphi) = 1/2 \,\E_{\hat P} \left[ ( \omega_y  - \left\langle  \omega  \right\rangle )^\top \, \bSigma \, ( \omega_y - \left\langle  \omega  \right\rangle ) \right] $.
This suggests an intriguing connection to variance-based regularization.
Weight-decay regularization, on the other hand, simply penalizes large norms.

\begin{algorithm}[t]
\caption{\small\,\,\textbf{Adversarial SGR.} Default values: $\lambda \in [0, 1]$,\, $\beta = 0.1$}
\begin{algorithmic}
\label{algorithm1}
\REQUIRE{Regularization strength $\lambda$ (noise-to-signal ratio), exponential decay rate $\beta$, batch size $m$}

\hspace{\algorithmicindent}{\,\,\,\,\,\, Average diagonal entry of data set variance-covariance matrix $c$ (constant for scaling)}

\hspace{\algorithmicindent}{\,\,\,\,\,\, Initial classifier parameters (weights and biases) $\omega_0$}
\WHILE{$\omega_t$ not converged}
	\STATE{Sample minibatch of data $\{(\x^{(1)}, y^{(1)}),...,(\x^{(m)}, y^{(m)}) \} \sim \hat{\P}$.}
	\STATE{Compute adversarial perturbations $\bxi^{(1)},...,\bxi^{(m)}$}
	\STATE{Compute covariance matrix or covariance function ${\rm Cov}(\bxi^{(1)},...,\bxi^{(m)})$ }
	\STATE{Update running average $\hat{\bSigma}_t \leftarrow (1\!-\!\beta)\, \hat{\bSigma}_{t-1} + \beta\, {\rm Cov}(\bxi^{(1)},...,\bxi^{(m)})$}
	\STATE{Compute scaling factor $\sigma_t = c / ( {\rm Avg} [\hat{\bSigma}_t]_{jj})$}
	\STATE{ \vspace{2mm} $\displaystyle \mathcal L(\omega) = \frac{1}{m}\, \sum \limits_{i=1}^m\ \Big[ \sum \limits_{k=1}^K - {y^{(i)}_k} \log \phi_{k}(\x^{(i)}; \omega) \Big]$}
	\STATE{ \vspace{2mm} $\displaystyle \Omega_{\hat{\bSigma}_t}(\omega) = \frac{\sigma_t}{2m}\, \sum \limits_{i=1}^m\ \bigg[ \sum \limits_{k=1}^K {y^{(i)}_k}\, \nabla_\x \log \phi_k(\x^{(i)}; \omega)^\top \, \hat{\bSigma}_t \, \nabla_\x \log \phi_k(\x^{(i)}; \omega) \bigg]$} \vspace{2mm}
	\STATE{$\displaystyle \omega_t \leftarrow \omega_{t-1} -  \left. \nabla_{\omega} \Big( \mathcal L(\omega) + \lambda \, \Omega_{\hat{\bSigma}_t}(\omega)  \Big) \right|_{\omega_{t-1}}$}
\ENDWHILE
\RETURN{$\omega_t$}
\end{algorithmic}
{The gradient-descent step can be performed with any learning rule. We used Adam in our experiments. See Sec.~\ref{sec:SGRimplementation} for an efficient implementation of the covariance gradient product.}
\end{algorithm}

\subsection{Implementation}
\vspace{-1mm}
\label{sec:SGRimplementation}
The matrix of raw second moments is aggregated through a running exponentially weighted average
\begin{align}
\hat{\bSigma}_t \leftarrow (1\!-\!\beta)\, \hat{\bSigma}_{t-1} + \beta\, {\rm Cov}(\bxi^{(1)},...,\bxi^{(m)})
\end{align}
with the decay rate $\beta$ as a tuneable parameter trading off weighting between current ($\beta\!\to\!1$) and past ($\beta\!\to\!0$) batch averages\footnote{We used $\beta\!=\!0.1$ in all our experiments.}. 
For low-dimensional data sets, we can directly compute the full matrix
\begin{align}
{\rm Cov}(\bxi^{(1)},...,\bxi^{(m)}) = {\frac 1 m} \sum_{i=1}^m \bxi_i \bxi_i^{\top} \, .
\label{eq:lowdim_cov}
\end{align}
For high-dimensional data sets, we have to resort to more memory-efficient representations leveraging the sparseness of the covariance matrix and covariance-gradient matrix-vector product.
For image data sets, the covariance matrix can be estimated as a function of the displacement between pixels, as illustrated in Fig.~\ref{fig:covariancefunctions},
\begin{align}
\left[{\rm Cov}(\bxi^{(1)},...,\bxi^{(m)}) \right]_{i\,j} \simeq {\rm CovFun}\left( || \mathbf{i} - \mathbf{j} ||_2 \right)
\end{align} 
where $\mathbf{i} = (i_{\rm r}, i_{\rm c})$ denotes the 2D pixel location, $i$ denotes the covariance matrix index obtained by flattening the 2D pixel location into a 1D row-major array such that $i_r = i \,\%\, {\rm image\_width}$, $i_c = i \,//\, {\rm image\_width}$ denote the row and column numbers of pixel $\mathbf{i}$ and similar for $\mathbf{j}$ respectively. 
Rounding pixel displacements $|| \mathbf{i} - \mathbf{j} ||_2$ to the nearest integer, the covariance function simply becomes an array of real numbers storing the current estimate of the covariance between two pixels displaced by the difference in array indices.
Note that it is not necessary to update $\hat{\bSigma}$ at every training step once a fairly good estimate is available.


In our experiments, we also standardized the covariance matrix such that the average diagonal entry is of the same order as the average diagonal entry of the data set covariance matrix. This allows for more intuitive optimization over the regularization strength parameter $\lambda$, which can now be interpreted as a noise-to-signal ratio.

The other crucial quantities needed to evaluate the regularizer are $\log\phi_{y_i}(\x_i)$ for training points $(\x_i,y_i)$. This is simply the per-sample cross-entropy loss, which is often available as a highly optimized callable operation in modern deep learning frameworks, reducing the implementation of the SGR regularizer to merely a few lines of code at very little computational overhead.


\section{Experiments}
\label{sec:experiments}

\subsection{Experimental Setup}

\textbf{Classifier architectures.}
We trained Convolutional Neural Networks (CNNs)~\cite{carlini2017towards} with nine hidden layers (and different numbers of filters for each data set), 
on CIFAR10~\cite{krizhevsky2009learning} and MNIST~\cite{lecun1998mnist}.
Our models are identical to those used in \cite{carlini2017towards, papernot2016distillation}. 
For both data sets, we adopt the following standard preprocessing and data augmentation scheme~\cite{resnet2016}: 
Each training image is zero-padded with four pixels on each side\footnote{We shifted the CIFAR10 data by $-0.5$ so that padding adds mid rgb-range pixels.}, randomly cropped to produce a new image with the original dimensions and subsequently horizontally flipped with probability one half. 
We also scale each image to have zero mean and unit variance when it is passed to the classifier.
Further details about the models, the learning rates and the optimizer can be found in Sec.~\ref{sec:ClassifierArchitecturesAppendix} in the Appendix.

\textbf{Training methods.}
We train each classifier with a number of different training methods: 
(i) clean, i.e. through the standard softmax cross-entropy objective, (ii) clean $+$ $L_2$ weight decay regularization, (iii) adversarially augmented training, i.e. training the classifier on a mixture of clean and adversarial examples (iv) GN (gradient-norm) regularized (corresponding to SGR with identity covariance matrix) and (v) SGR regularized. The SGR covariance matrix was computed from either: (a) loss gradients, (b) sign of loss gradients, (c) FGSM or (d) PGD perturbations. The performance of these SGR variants was almost identical in our experiments.

\textbf{Adversarial attacks.}
We evaluate the classification accuracy against the following adversarial attacks:
Fast Gradient (Sign) Method (FGM \& FGSM)~\cite{goodfellow2014explaining}, Projected Gradient Descent (PGD)~\cite{madry2017towards} and DeepFool~\cite{moosavi2016deepfool}. 
The attack strength $\epsilon$ is reported in units of $1/255$. 
The numbers reported for the DeepFool attack are computed according to Eq.~(2) in \cite{moosavi2016deepfool}.
The number of iterations are $10$ / $40$ (white-box / transfer attack tables) for PGD and $100$ for DeepFool.
The attacks were implemented with the open source CleverHans Library \cite{papernot2017cleverhans}.
We also implemented a uniform random noise baseline.
Further details and attack hyper-parameters can be found in Sec.~\ref{sec:AttackHyperparamsAppendix} the Appendix.

\begin{figure}[t]
\begin{center}
\begin{tabular}{@{}c@{}c@{}c@{}c@{}c@{}}
{\setlength{\fboxsep}{0pt}\fbox{\adjincludegraphics[width=0.19\columnwidth , trim={{0.2\width} {0.25\height} {0.384\width} {0.25\height}},clip]{/CovarianceMatrices/cifar10_clean/{Sigma_pgd_attack_TrueIntensity_bwr}.png}}} &
\hspace{6pt}{\setlength{\fboxsep}{0pt}\fbox{\adjincludegraphics[width=0.19\columnwidth , trim={{0.2\width} {0.25\height} {0.384\width} {0.25\height}},clip]{/CovarianceMatrices/cifar10_clean/{Sigma_fgsm_attack_TrueIntensity_bwr}.png}}} &
\hspace{6pt}{\setlength{\fboxsep}{0pt}\fbox{\adjincludegraphics[width=0.19\columnwidth , trim={{0.2\width} {0.25\height} {0.384\width} {0.25\height}},clip]{/CovarianceMatrices/cifar10_clean/{Sigma_fool_attack_TrueIntensity_bwr}.png}}} &
\hspace{6pt}{\setlength{\fboxsep}{0pt}\fbox{\adjincludegraphics[width=0.19\columnwidth , trim={{0.2\width} {0.25\height} {0.384\width} {0.25\height}},clip]{/{LongRangeCorrelated}/{CIFAR10_Cov_Natural}.png}}} &
\hspace{6pt}\adjincludegraphics[height=2.7cm, trim={{0.\width} {0.099\height} {0.\width} {0.09\height}},clip]{/colorbar.png}
\end{tabular}
\vspace{-1mm}
\caption{Covariance matrices of PGD, FGSM and DeepFool perturbations as well as CIFAR10 training set (for comparison). The short-range structure of the perturbations is clearly visible. It is also apparent that the first two attack methods yield perturbations with almost identical  covariance structure. We are showing a center-crop for better visibility (25\% trimmed on each side). For comparison the matrices were rescaled such that their maximum element has an absolute value of~one. }
\label{fig:adversarialcovariance}
\vspace{2mm}
\begin{tabular}{@{}c@{\hspace{10pt}}c@{}}
\adjincludegraphics[width=0.43\linewidth, trim={{0} {0} {0} {0.5\height}},clip]{/{CovarianceFunction/pgd_clean}/{pgd_Horizontal_IntraChannel_CovarianceFunction_row15_col15}.pdf} &
\adjincludegraphics[width=0.43\linewidth, trim={{0} {0} {0} {0.5\height}},clip]{/{CovarianceFunction/pgd_clean}/{pgd_Horizontal_InterChannel_CovarianceFunction_row15_col15}.pdf} 
\end{tabular}
\vspace{-3mm}
\caption{
PGD covariance functions: (Left) intra-channel covariance functions, measuring correlations between identical color channels, (the coloring in the plot matches the corresponding channel), (right) inter-channel covariance functions, measuring correlations between opposite color channels. The rapid decay with pixel displacement, indicating very short decay length, is clearly visible.}
\label{fig:covariancefunctions}
\end{center}
\vspace{-1mm}
\end{figure}

\subsection{Covariance Structure of Adversarial Perturbations} 
\label{sec:covstructureofadvperts}
To investigate the covariance structure of adversarial perturbations, 
we trained a clean classifier for $50$ epochs and computed adversarial perturbations for every data point in the test set.
The 2D perturbations were flattened into 1D row-major arrays before computing the full covariance matrix\footnote{Strictly speaking, $\hat{\bSigma}$ denotes the matrix of raw second moments. We will refer to $\hat{\bSigma}$ as the covariance matrix however, since the difference between the two is safely negligible, with the raw second moments typically being three orders of magnitude larger than the outer-product of the mean perturbations.} as described in Section~\ref{sec:SGRimplementation}, \Eq{\ref{eq:lowdim_cov}}. 

The short-range correlation structure of the perturbations, shown in Fig.~\ref{fig:adversarialcovariance}, is clearly visible (the correlations decay much faster than those of the data set covariance matrix).
Figure~\ref{fig:covariancefunctions} shows the corresponding PGD covariance functions.
The rapid decay with pixel displacement, indicating very short decay length\footnote{Decay length is defined as the displacement over which the covariance function decays to $1/e$ of its value.}, is again evident.
It thus seems that an unregularized classifier vulnerable to adversarial perturbations gives too much weight to short-range correlations (low-level patterns) and not enough weight to long-range ones (globally relevant high-level features of the data).

\subsection{Long-range Correlated Noise Attack}
\label{sec:LRCattack}
To investigate the effect and potential benefit of using a structured covariance matrix in the SGR regularizer versus an ``unstructured'' diagonal covariance, corresponding to simple gradient-norm regularization, 
we sample perturbations from a long-range correlated multivariate Gaussian 
with covariance matrix specified through an exponentially decaying covariance function parametrized by a variable decay length $\zeta$.
Inspired by the PGD covariance function, we chose intra-channel CovFun $f(r) = \exp(-r / \zeta)$ and inter-channel CovFun $f( r ) = 0.5 \exp(-r / \zeta)$.
The corresponding covariance matrices are depicted in Fig.~\ref{fig:longrangecovariances} in the Appendix. LRC samples are shown in Fig.~\ref{fig:LRCnoise}.

The worst-of-$100$ attack then consists in perturbing every test set data point $100$ times and evaluating the classifier accuracy against the worst of these perturbations. 
We tested for several decay lengths in the range $\zeta \in [1,2,4,8,16]$ and attack strengths $\epsilon \in [0.01, 0.7]$ with which the perturbations were scaled.
Figure~\ref{fig:LRCnoiseAppendix} in the Appendix shows the accuracy of different models as a function of $\epsilon$.
As a baseline, we also trained a model on LRC-augmented input. For SGR and GN we performed a hyper-parameter search over $\lambda$ (at fixed $\zeta = 8$) and report results for the best performing model,
which we found to be $\lambda\!=\!5.0$ for SGR and $\lambda\!=\!0.1$ for GN.

As can be seen in Fig.~\ref{fig:LRCnoise}, SGR clearly and consistently outperforms GN on long-range correlated perturbations, achieving over $25\%$ higher classification accuracy on LRC-perturbed input with $\zeta\sim16$.
As the decay length goes to zero, the synthetic covariance matrix converges to the identity matrix and SGR performance approaches GN performance. 

Finally, to verify that those effects are not a result of our synthetically chosen covariance function, we also performed the same experiments with the covariance matrix computed from the CIFAR10 training set, shown in Fig.~\ref{fig:adversarialcovariance}.
The accuracies for $\epsilon=0.3$ are: SGR $65.8\%$, GN $48.7\%$, LRC noise $57.9\%$ and clean $46.5\%$,
again confirming the benefit of using a structured covariance in the regularizer.

\begin{figure}[t]
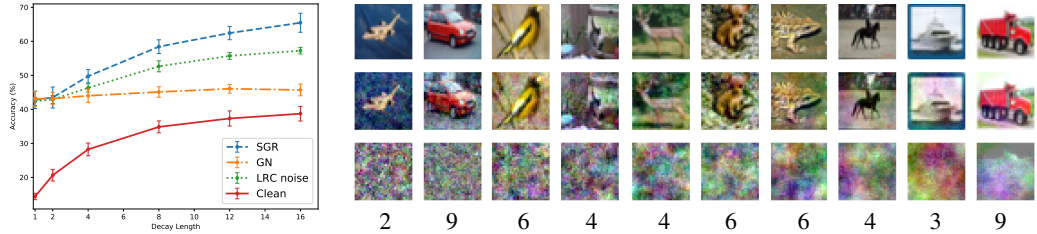

\begin{center}
\begin{tabular}{@{\hspace{0pt}}c@{\hspace{3pt}}c@{\hspace{4.5pt}}c@{\hspace{4pt}}c@{\hspace{3pt}}c@{\hspace{0pt}}c@{}}
\hspace{-1mm}\multirow{2}{*}[2.57cm]{\adjincludegraphics[width=0.31\linewidth]{/LongRangeCorrelated/{Avg_Worst-of-100_LRC-noise_attack_errorbars}.pdf} }\,\,
&
\adjincludegraphics[width=0.125\linewidth, trim={{0.12\width} {0.29\height} {0.73\width} {0.13\height}},clip]{/{LongRangeCorrelated/samples}/{grid_lrn_2018_05_12_0643_cifar10_clean_0.001lnr_4.0eps_1.0lrndecaylength_0.3lrnnsr}.png} & \hspace{-4pt}
\adjincludegraphics[width=0.125\linewidth, trim={{0.28\width} {0.29\height} {0.57\width} {0.13\height}},clip]{/{LongRangeCorrelated/samples}/{grid_lrn_2018_05_12_0215_cifar10_clean_0.001lnr_4.0eps_2.0lrndecaylength_0.3lrnnsr}.png} & 
\adjincludegraphics[width=0.125\linewidth, trim={{0.44\width} {0.29\height} {0.41\width} {0.13\height}},clip]{/{LongRangeCorrelated/samples}/{grid_lrn_2018_05_11_2205_cifar10_clean_0.001lnr_4.0eps_4.0lrndecaylength_0.3lrnnsr}.png} &
\adjincludegraphics[width=0.125\linewidth, trim={{0.6\width} {0.29\height} {0.25\width} {0.13\height}},clip]{/{LongRangeCorrelated/samples}/{grid_lrn_2018_05_11_1758_cifar10_clean_0.001lnr_4.0eps_8.0lrndecaylength_0.3lrnnsr}.png} &
\adjincludegraphics[width=0.125\linewidth, trim={{0.75\width} {0.29\height} {0.10\width} {0.13\height}},clip]{/{LongRangeCorrelated/samples}/{grid_lrn_2018_05_11_1351_cifar10_clean_0.001lnr_4.0eps_12.0lrndecaylength_0.3lrnnsr}.png} \\[-1mm]
& \,{\small 2} \hspace{0.555cm} {\small 9} & {\small 6} \hspace{0.57cm} {\small 4} & {\small 4} \hspace{0.55cm} {\small 6} & {\small 6} \hspace{0.55cm} {\small 4} & {\small 3} \hspace{0.53cm} {\small 9} \\
\end{tabular} 
\vspace{-0.5mm}
\caption{
(Left) Accuracy of different models as a function of the LRC attack decay length $\zeta \in [1,16]$ at fixed $\epsilon = 0.3$ (averaged over five runs).
The plot clearly demonstrates the superiority of using a structured covariance matrix in the SGR regularizer.
(Right) Pairs of long range correlated noise perturbed samples with increasing decay lengths $\zeta \in [1,2,4,8,12]$. 
Top row shows original test set images, middle shows adversarially perturbed samples, while the bottom row shows
the adversarial perturbations rescaled to fit the full range of rgb values (their actual range and hence saliency is smaller).
The footer indicates the classifier predictions on adversarial input.}
\label{fig:LRCnoise}
\end{center}
\end{figure}

\subsection{White-box \& Transfer Attack Accuracy}
\label{sec:whiteboxandtransferattacks}
A summary of the white-box\footnote{In the white-box setting the attacker has full access to the model and is in particular able to backpropagate through it to find adversarial perturbations.} and transfer attack accuracies  
can be found in Table~\ref{table:whiteboxandtransferaccuracies}.
As for the LRC experiment, we performed a hyper-parameter search over the SGR/GN regularization strength~$\lambda$. 
Our aim was to get the highest possible white-box accuracy \textit{without compromising on the test set accuracy.}\footnote{This is in line with the notion that a regularizer should help the model to become more robust and not completely alter its original learning task.} 
On MNIST for an attack strength of $\epsilon\!=\!32$, SGR \& GN achieve remarkable PGD white-box attack accuracies of $90\%$ compared to the $44.8\%$ accuracy achieved by the clean model. 
On CIFAR10 for $\epsilon\!=\!8$, SGR \& GN did not achieve such high white-box numbers without trading-off test set accuracy (see non-bold diagonal entries). 
This is in contrast to the remarkably high white-box accuracy achieved by the PGD-augmented model, which however comes at the expense of a considerable drop in test set accuracy.

To investigate whether the high white-box accuracy achieved by the PDG-adversarially trained model is due to improved robustness or whether it is in part due to adversarial training resulting in PGD producing adversarial perturbations that become easier to classify\footnote{As hypothesized in \cite{galloway2018adversarial}, it could be that adversarial training incentivises the classifier to deliberately contort its decision boundaries such that adversarial perturbations become easier to classify.}, we henceforth evaluated transfer attack accuracies.
The idea of transfer attacks is to swap adversarial samples generated by models trained through different training methods.
The PGD\footnote{PGD here refers to the attack, not to be confused with the PGD-augmented adversarially trained model. It is always clear from the context which one is meant.} transfer attack accuracies 
were evaluated on the full test set (see bold-face off-diagonal entries).
All SGR variants performed equally well. We show results for SGR-sign, see Tables~\ref{fig:cifarwhiteboxappendix}\,\&\,\ref{fig:transferappendix} in the Appendix for other SGR variants.
Examples of PGD perturbed samples can be found in Figs.~\ref{fig:PGDmnistamples}\,\&\,\ref{fig:PGDcifarsamples} in the Appendix.

An interesting observation can be made from the transfer attack accuracies of the clean model, shown in the first row of the table.
As can be seen, the PGD adversarial perturbations obtained from SGR/GN trained models seem to hurt the clean model much more than the PGD perturbations obtained from the PGD/FGSM adversarially trained models. This could be due to either (i) the SGR/GN induced adversarial perturbations being ``stronger'' or (ii) the SGR/GN regularized models being similar to the clean one, such that the transfer attack acts more like a white-box attack. Both alternatives are good news for the regularized variants in a way. Looking at the column for PGD-adversarial samples generated from the clean trained model, it however appears that those perturbations do not equally hurt the SGR/GN regularized models, thus rather supporting hypothesis~(i).
We leave it to the reader to assess the remaining numbers in favour of or against this conjecture.

We have already discussed the potential benefits of SGR over its simpler counterpart GN in the case of long-range correlated perturbations. As expected, the difference between SGR and GN in terms of white-box and transfer attack accuracies is less significant in this experiment,
which can be explained by the evident short-range nature of current adversarial attacks (see discussion in Sec.~\ref{sec:covstructureofadvperts}). Reassuringly, we did not observe any loss of performance in all our experiments when using the covariance function instead of the full covariance matrix in the SGR regularizer.

\begin{table*}
\begin{tabular}{ c c }
{\small MNIST $\epsilon=32$} & {\small CIFAR10 $\epsilon=8$ }\\[1mm]
 \hspace{-4mm}
{\tiny
\begin{tabular}{l c  c  c  c  c  c }
 \\
\textsc{Train.\,Meth.} & \hspace{-5pt}\textsc{test}\hspace{-5pt} & \hspace{-3pt}\textsc{rand}\hspace{-3pt} & \textsc{fgm} & \hspace{-3pt}\textsc{fgsm}\hspace{-3pt} & \textsc{pgd} & \textsc{fool} \\

\hline
\textsc{clean}  				& {\bf 99.4} & {\bf 99.4} & {\bf 98.0} & {\bf 89.1} & {\bf 44.8} & {\bf 0.159} \\
\textsc{wdecay} 			& {\bf 99.4} & {\bf 99.3} & {\bf 99.4} & {\bf 88.2} & {\bf 59.8} & {\bf 0.209} \\
\textsc{pgd} 				& {\bf 99.5} & {\bf 99.5} & {\bf 99.5} & {\bf 98.8} & {\bf 98.3} & {\bf 0.423} \\
\textsc{fgsm} 				& {\bf 99.4} & {\bf 99.4} & {\bf 99.2} & {\bf 98.5} & {\bf 96.9} & {\bf 0.325} \\
\textbf{\textsc{gn}}  			& {\bf 99.4} & {\bf 99.3} & {\bf 99.3} & {\bf 94.2} & {\bf 90.0} & {\bf 0.307} \\
\textbf{\textsc{sgr}} 			& {\bf 99.5} & {\bf 99.5} & {\bf 99.5} & {\bf 94.9} & {\bf 89.5} & {\bf 0.301} \\
\hline
\end{tabular} }
&\hspace{-3mm}
{\tiny
\begin{tabular}{l  c  c  c  c  c  c  c } 
& & \multicolumn{6}{c}{ \textsc{Train.\,Meth.} } \\
\textsc{Train.\,Meth.} & \hspace{-3pt}{\textsc{test}}\hspace{-3pt} & \hspace{-5pt}\textsc{clean}\hspace{-5pt} & \textsc{wd} & \textsc{pgd} & \hspace{-3pt}\textsc{fgsm}\hspace{-3pt} & \textsc{gn} & \textsc{sgr}  \\
\hline
\textsc{clean} 					&{\bf 84.3} & 11.2 & {\bf 27.7} & {\bf 61.6} & {\bf 68.1} & {\bf 27.8} & {\bf 26.5} \\
\textsc{wdecay} 				&{\bf 85.2} & {\bf 44.5} & 8.0 & {\bf 63.7} & {\bf 68.6} & {\bf 25.6} & {\bf 24.2} \\
\textsc{pgd} 					&{\bf 77.7} & {\bf 76.2} & {\bf 75.0} & 40.5 & {\bf 71.3} & {\bf 66.6} & {\bf 67.3} \\
\textsc{fgsm}					&{\bf 81.2} & {\bf 75.9} & {\bf 74.0} & {\bf 60.3} & 13.2 & {\bf 62.9} & {\bf 63.8} \\
\textbf{\textsc{gn}}  				&{\bf 83.3} & {\bf 75.0} & {\bf 70.1} & {\bf 63.9} & {\bf 70.3} & 11.2 & {\bf 44.0} \\
\textbf{\textsc{sgr}} 				&{\bf 84.2} & {\bf 74.3} & {\bf 67.8} & {\bf 65.3} & {\bf 70.9} & {\bf 40.5} & 10.0 \\
\hline
\end{tabular} }
\end{tabular}
\caption{(Left): Test set and white-box attack accuracies.
(Right): Test set accuracies as well as white-box attack (non-bold diagonal entries) and PGD-transfer attack (bold-face off-diagonal entries) accuracies. 
The corresponding numbers for the other data set, as well as SGR/GN hyper-parameters can be found in Tables~\ref{fig:cifarwhiteboxappendix}\,\&\,\ref{fig:transferappendix} in the Appendix. The results are discussed in Sec.~\ref{sec:whiteboxandtransferattacks}.}
\label{table:whiteboxandtransferaccuracies}
\end{table*}


\section{Related Work}
\label{sec:relatedwork}

A large body of related work on distributionally robust optimization~\cite{sinha2017certifiable, gao2016distributionally} seeks a classifier that minimizes the worst-case loss over a family of distributions that are close to the empirical distribution in terms of $f$-divergence or Wasserstein distance.
While these methods aim at identifying the worst-case distribution,
we propose to efficiently approximate expectations over corrupted distributions through the use of a regularizer informed by adversarial perturbations, which can themselves be understood as finite budget approximations to worst-case corruptions.

Coincidentally, \cite{simon2018adversarial} also suggested to use gradient-norm regularization to improve the robustness against adversarial attacks.
While they investigated the effect of different gradient-norms induced by constraints on the magnitude of the perturbations, we focused on a data-dependent \textit{structured} generalization of gradient regularization.

Much work has also been done on the connection between overfitting, generalization performance and representational power of neural networks~\cite{delalleau2011shallow, eldan2016power, cohen2016convolutional, zhang2016understanding}.
Similar in spirit to our work is also the recent work on adversarial training vs.\ weight decay regularization~\cite{galloway2018adversarial}. 
While weight decay acts on the parameters, our regularizer acts on the function realized by the classifier. As discussed in Sec.~\ref{sec:SGRproperties}, the two approaches are complementary to each other and can also be combined.

From a Bayesian perspective, regularization can be understood as imposing a prior distribution over the model parameters. 
As parameters in neural networks are non-identifiable, we would however ultimately prefer to impose priors in the space of functions directly~\cite{yeewhye2017bayesiandeeplearning, neal2012bayesian, neal1996priors}. 
As pointed out in Sec.~\ref{sec:SGRproperties}, this is indeed one of the key properties of our regularizer.


\section{Conclusion}
\label{sec:conclusion}

The fact that adversarial perturbations can fool classifiers while being imperceptible to the human eye, hints at a fundamental mismatch between how humans and classifiers try to make sense of the data they see.
We provided evidence that current adversarial attacks act by perturbing the short-range correlations of signals.
We proposed a novel \textit{structured gradient regularizer} (SGR) informed by the covariance structure of adversarial noise and
presented an efficient SGR implementation with low memory footprint.
Our results suggest a flurry of future work, 
augmenting the theoretical and experimental aspects of attack specific adversarial examples to a more general robustness against structured perturbations.

\section*{Acknowledgements}
KR would like to thank Eirikur Agustsson, Yannic Kilcher, Andreas Marfurt and Jonas Kohler for insightful discussions.
This work was supported by Microsoft Research through its PhD Scholarship Programme. 

\bibliography{nips_2018}
\bibliographystyle{plain}


\newpage
\section{Appendix}
\label{sec:appendix}

\subsection{SGR properties}
\label{sec:SGRpropertiesAppendix}

(iii) We can gain complementary insights into SGR by explicitly computing it in terms of  logits 
\begin{align}
\phi_y(\x) = \frac{\exp{( \varphi_y(\x) )}}{\sum_y \exp{( \varphi_y(\x) ) } } \,
\end{align} 
Defining the 
class average of the logit-gradients as 
\begin{align}
\left\langle \nabla \varphi  \right\rangle(\x) := \sum_y \nabla \varphi_y(\x) \phi_y(\x)
\label{eq:thermodynamicmeanweight}
\end{align} 
we obtain the following expression for the structured gradient regularizer
\begin{align}
\hspace{-3pt}\Omega_{\bSigma}(\bphi) = \frac{1}{2} \E_{\hat P} \bigg[ ( \nabla\varphi_y \hspace{-1pt}-\hspace{-1pt} \left\langle  \nabla\varphi  \right\rangle )^\top \bSigma ( \nabla\varphi_y \hspace{-1pt}-\hspace{-1pt} \left\langle  \nabla\varphi  \right\rangle ) \bigg]
\end{align}
We can therefore see that SGR is penalizing large \textit{fluctuations} of the class-conditional logit-gradients $\nabla\varphi_y$ around their data-dependent class average $\left\langle  \nabla\varphi  \right\rangle$. 
For simple one-layer softmax classifiers $\varphi_y(\x) = \omega_y^\top \x + b_y$, we obtain
\begin{align}
\Omega_{\bSigma}(\bphi) &= \frac{1}{2} \E_{\hat P} \bigg[ ( \omega_y  - \left\langle  \omega  \right\rangle )^\top \, \bSigma \, ( \omega_y - \left\langle  \omega  \right\rangle ) \bigg]\,
\end{align} 
This suggests an intriguing connection to variance-based regularization.
Weight-decay regularization, on the other hand, simply penalizes large norms.

\subsection{Attack Hyper-parameters}
\label{sec:AttackHyperparamsAppendix}
An overview of the attack hyper-parameters can be found in Table~\ref{table:attachhyperparams_}.
All of the attacks are untargeted, meaning that they find adversarial perturbations to an arbitrary one of the $K\!-\!1$ other classes. 
Both PGD and DeepFool are iterative attacks: they use $10$ / $40$ (for the white-box / transfer attack tables) and $100$ iterations respectively. 
The numbers reported for the DeepFool attack are computed according to Eq.~(2) in \cite{moosavi2016deepfool}.
The adversarial example construction processes use the most likely label predicted by the classifier in order to avoid label leaking \cite{kurakin2016adversarialatscale} during adversarially augmented training.
The perturbed images returned by the attacks were clipped to lie within the original rgb range, as is commonly done in practice.
The attacks were implemented with the open source CleverHans Library~\cite{papernot2017cleverhans}.

\begin{table}[h]
\begin{center}
\begin{tabular}{l l}
Attack & Hyper-parameters \\[1mm]
\hline\\[-2mm]
\textsc{rand} & $\epsilon$ : in units of $1/255$ \\[2mm]

\textsc{fgsm} & $\epsilon$ : in units of $1/255$ \\[2mm]

\textsc{pgd} & $\epsilon$ : in units of $1/255$ \\
& $\epsilon_{\rm iter} = \epsilon / 5$ \\
& ${\rm nb\_iter} = 10 / 40$ (white-box / transfer attack tables) \\[2mm]

\textsc{deep fool} & overshoot : $0.02$ \\
& max\_iter$ = 100$ \\[1mm]
\hline
\end{tabular}
\vspace{5mm}
\caption{Attack hyper-parameters. }
\label{table:attachhyperparams_}
\end{center}
\end{table}

\newpage

\subsection{Further Experimental Results}

\vspace{.5cm}

\begin{figure}[h]
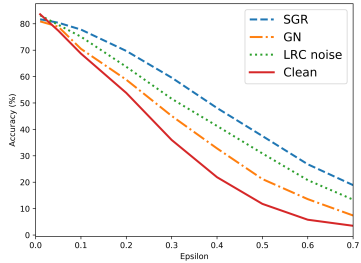

\begin{center}
\adjincludegraphics[width=0.35\linewidth]{/LongRangeCorrelated/{Worst-of-100_LRC-noise_attack_eps}.png}
\caption{
Long Range Correlated Noise. Classification accuracy of different models as a function of the attack strength $\epsilon$ for an LRC attack with a fixed decay length $\zeta = 8$ on CIFAR10. The SGR/GN hyper-parameters are reported in Sec.~\ref{sec:LRCattack}. See Fig.~\ref{fig:LRCnoise} for a plot of the accuracy of different models as a function of the LRC attack decay length $\zeta \in [1,16]$ at fixed $\epsilon = 0.3$.}
\label{fig:LRCnoiseAppendix}
\end{center}
\end{figure}

\begin{figure*}[h]
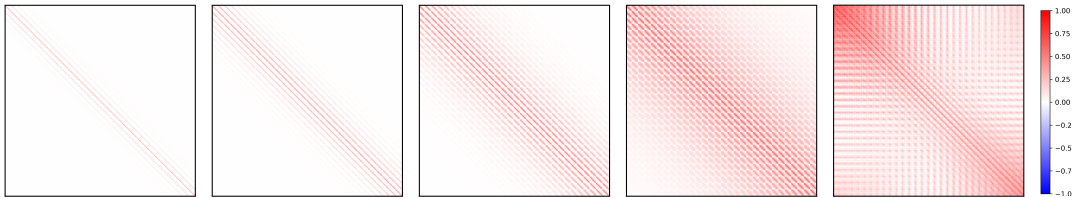

\begin{center}
\begin{tabular}{@{}c@{}c@{}c@{}c@{}c@{}c@{}}
{\setlength{\fboxsep}{0pt}\fbox{\adjincludegraphics[width=0.18\linewidth, trim={{0.05\width} {0.07\height} {0.24\width} {0.07\height}},clip]{/{LongRangeCorrelated}/{CIFAR10_Cov_DecayLength1_700dpi}.png}}} & 
\hspace{6pt}{\setlength{\fboxsep}{0pt}\fbox{\adjincludegraphics[width=0.18\linewidth, trim={{0.05\width} {0.07\height} {0.24\width} {0.07\height}},clip]{/{LongRangeCorrelated}/{CIFAR10_Cov_DecayLength2_700dpi}.png}}} &
\hspace{6pt}{\setlength{\fboxsep}{0pt}\fbox{\adjincludegraphics[width=0.18\linewidth, trim={{0.05\width} {0.07\height} {0.24\width} {0.07\height}},clip]{/{LongRangeCorrelated}/{CIFAR10_Cov_DecayLength4_700dpi}.png}}} &
\hspace{6pt}{\setlength{\fboxsep}{0pt}\fbox{\adjincludegraphics[width=0.18\linewidth, trim={{0.05\width} {0.07\height} {0.24\width} {0.07\height}},clip]{/{LongRangeCorrelated}/{CIFAR10_Cov_DecayLength8_700dpi}.png}}} &
\hspace{6pt}{\setlength{\fboxsep}{0pt}\fbox{\adjincludegraphics[width=0.18\linewidth, trim={{0.05\width} {0.07\height} {0.24\width} {0.07\height}},clip]{/{LongRangeCorrelated}/{CIFAR10_Cov_Natural}.png}}} &
\hspace{6pt}\adjincludegraphics[height=2.55cm, trim={{0.\width} {0.099\height} {0.\width} {0.09\height}},clip]{/colorbar.png}
\end{tabular}
\caption{
Long-range structured covariance matrices corresponding to the covariance functions defined in Sec.~\ref{sec:LRCattack} for increasing decay lengths $\zeta \in [1, 2, 4, 8]$, as well as covariance matrix of CIFAR10 training set (for comparison).
As an alternative to the covariance function representation described in Sec.~\ref{sec:SGRimplementation}, we can also leverage the block-structure of the full covariance matrix to obtain a sparse representation:
as can be seen in the figures, different equally distant pairs of rows are often nearly identically correlated. Hence it is enough to estimate the covariance between a few pairs of rows $r_m$ and $r_{n} > r_m$ (until the correlations become sufficiently small). This amounts to for instance estimating the covariance matrix by a diagonal and a few off-diagonal blocks.}
\label{fig:longrangecovariances}
\end{center}
\end{figure*}

\begin{figure*}[h]
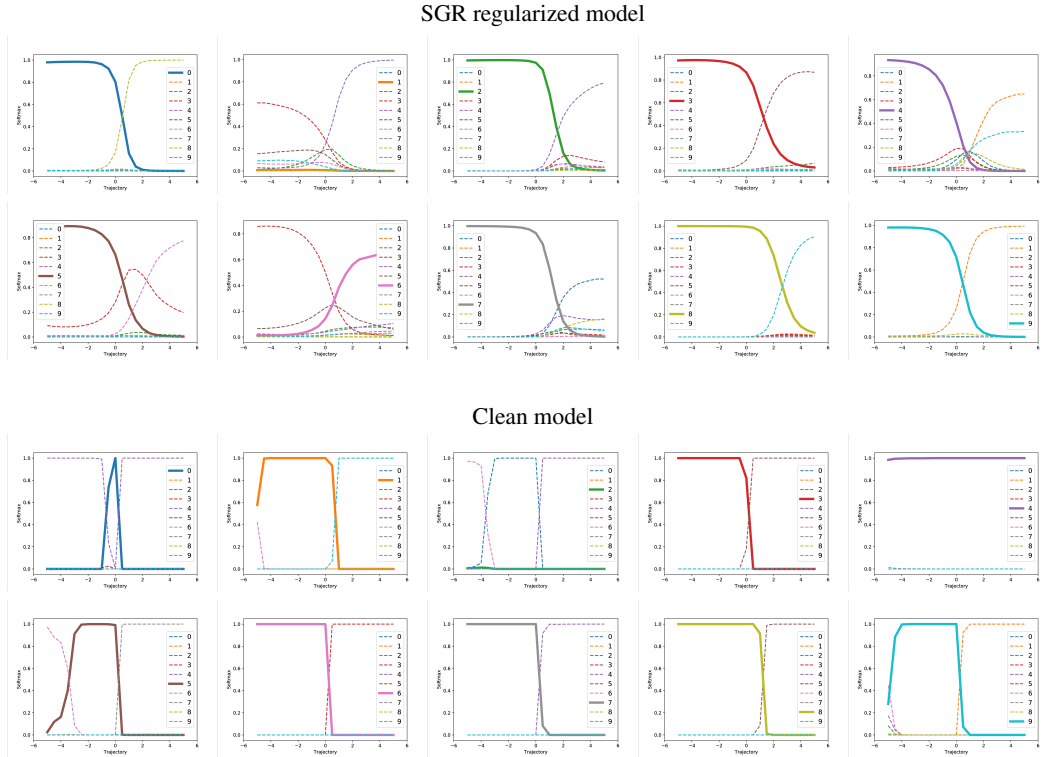

  \begin{center}
  {\small SGR regularized model}\\[1mm]
    \begin{tabular}{@{}c@{}c@{}c@{}c@{}c@{}}
    \adjincludegraphics[width=0.2\linewidth]{/LinearExtrapolation/cifar10_sgr/{linear_extrapolation_pgd_class0}.png} &
    \adjincludegraphics[width=0.2\linewidth]{/LinearExtrapolation/cifar10_sgr/{linear_extrapolation_pgd_class1}.png}  &
    \adjincludegraphics[width=0.2\linewidth]{/LinearExtrapolation/cifar10_sgr/{linear_extrapolation_pgd_class2}.png}  &
    \adjincludegraphics[width=0.2\linewidth]{/LinearExtrapolation/cifar10_sgr/{linear_extrapolation_pgd_class3}.png}  &
    \adjincludegraphics[width=0.2\linewidth]{/LinearExtrapolation/cifar10_sgr/{linear_extrapolation_pgd_class4}.png}  \\
    \adjincludegraphics[width=0.2\linewidth]{/LinearExtrapolation/cifar10_sgr/{linear_extrapolation_pgd_class5}.png}  &
    \adjincludegraphics[width=0.2\linewidth]{/LinearExtrapolation/cifar10_sgr/{linear_extrapolation_pgd_class6}.png}  &
    \adjincludegraphics[width=0.2\linewidth]{/LinearExtrapolation/cifar10_sgr/{linear_extrapolation_pgd_class7}.png}  &
    \adjincludegraphics[width=0.2\linewidth]{/LinearExtrapolation/cifar10_sgr/{linear_extrapolation_pgd_class8}.png}  &
    \adjincludegraphics[width=0.2\linewidth]{/LinearExtrapolation/cifar10_sgr/{linear_extrapolation_pgd_class9}.png}   
    \end{tabular}
    \\[5mm]
    {\small Clean model}\\[1mm]
    \begin{tabular}{@{}c@{}c@{}c@{}c@{}c@{}}
    \adjincludegraphics[width=0.2\linewidth]{/LinearExtrapolation/cifar10_clean/{linear_extrapolation_pgd_class0}.png} &
    \adjincludegraphics[width=0.2\linewidth]{/LinearExtrapolation/cifar10_clean/{linear_extrapolation_pgd_class1}.png}  &
    \adjincludegraphics[width=0.2\linewidth]{/LinearExtrapolation/cifar10_clean/{linear_extrapolation_pgd_class2}.png}  &
    \adjincludegraphics[width=0.2\linewidth]{/LinearExtrapolation/cifar10_clean/{linear_extrapolation_pgd_class3}.png}  &
    \adjincludegraphics[width=0.2\linewidth]{/LinearExtrapolation/cifar10_clean/{linear_extrapolation_pgd_class4}.png}  \\
    \adjincludegraphics[width=0.2\linewidth]{/LinearExtrapolation/cifar10_clean/{linear_extrapolation_pgd_class5}.png}  &
    \adjincludegraphics[width=0.2\linewidth]{/LinearExtrapolation/cifar10_clean/{linear_extrapolation_pgd_class6}.png}  &
    \adjincludegraphics[width=0.2\linewidth]{/LinearExtrapolation/cifar10_clean/{linear_extrapolation_pgd_class7}.png}  &
    \adjincludegraphics[width=0.2\linewidth]{/LinearExtrapolation/cifar10_clean/{linear_extrapolation_pgd_class8}.png}  &
    \adjincludegraphics[width=0.2\linewidth]{/LinearExtrapolation/cifar10_clean/{linear_extrapolation_pgd_class9}.png}   
    \end{tabular}
    
    \caption{
    Softmax activations along a linear trajectory between a clean sample and plus/minus five times the $\epsilon=4$ PGD adversarial perturbation for an SGR regularized classifier (top) and a clean classifier (bottom) on CIFAR10.}
    \label{fig:softmaxtrajectoriesappendix}
  \end{center}
\end{figure*}

\vspace{.5cm}
\paragraph{Smoothness of Regularized Classifier}
At the most basic level of abstraction, adversarial vulnerability is related to an overly-steep classifier gradient, such that a tiny change in the input can lead to a large perturbation of the classifier output.
In order to analyze this effect, we visualize the softmax activations along linear trajectories between various clean and corresponding adversarial examples. As can be seen in Fig.~\ref{fig:softmaxtrajectoriesappendix}, the regularized classifier consistently leads to smooth classifier outputs.

\vspace{.5cm}
\paragraph{Label leaking.}

Label leaking can cause models trained with adversarial examples to perform better on perturbed than on clean inputs, as the model can learn to exploit regularities in the adversarial example construction process~\cite{kurakin2016adversarialatscale}. This hints at a danger of overfitting to the specific attack when training with adversarial examples. SGR regularized classifiers, on the other hand, do not suffer from label leaking, as the adversarial examples are never directly fed to the classifier.

\begin{table*}[h!]
\begin{center}
\begin{tabular}{ c c }
{\small MNIST $\epsilon=32$} & {\small CIFAR10 $\epsilon=4$ }\\[1mm]
{\tiny
\hspace{-5mm}\begin{tabular}{l  c  c  c  c  c  c }
{\textsc{train.\,meth}} & \hspace{-5pt}\textsc{test}\hspace{-5pt} & \hspace{-3pt}\textsc{rand}\hspace{-3pt} & \textsc{fgm} & \hspace{-3pt}\textsc{fgsm}\hspace{-3pt} & \textsc{pgd} & \textsc{fool} \\
\hline
\textsc{clean}					 		& 99.4 & 99.4 & 98.0 & 89.1 & 44.8 & 0.159 \\
\textsc{wdecay} $2e\!-\!4$			 		& 99.4 & 99.3 & 99.4 & 88.2 & 59.8 & 0.209 \\
\textsc{pgd}		 			 		& 99.5 & 99.5 & 99.5 & 98.8 & 98.3 & 0.423 \\
\textsc{fgsm}					 		& 99.4 & 99.4 & 99.2 & 98.5 & 96.9 & 0.325 \\
\textsc{gn} $\lambda\!=\!5.0$ 		 		& 99.4 & 99.3 & 99.3 & 94.2 & 90.0 & 0.307 \\
\textsc{sgr\_grad} $\lambda\!=\!0.1$ 	 		& 99.5 & 99.5 & 99.5 & 94.3 & 88.5 & 0.294 \\
\textsc{sgr\_sign} $\lambda\!=\!1.0$	 		& 99.5 & 99.5 & 99.5 & 94.9 & 89.5 & 0.301 \\
\textsc{fgsm\_sgr} $\lambda\!=\!0.5$	 		& 99.5 & 99.5 & 99.5 & 94.0 & 83.2 & 0.284 \\
\textsc{pgd\_sgr} $\lambda\!=\!1.0$	 		& 99.6 & 99.5 & 99.5 & 94.4 & 84.3 & 0.294 \\
\hline
\end{tabular} }
&\hspace{-3mm}
{\tiny
\begin{tabular}{l  c  c  c  c  c  c }
{\textsc{train.\,meth}} & \hspace{-5pt}\textsc{test}\hspace{-5pt} & \hspace{-3pt}\textsc{rand}\hspace{-3pt} & \textsc{fgm} & \hspace{-3pt}\textsc{fgsm}\hspace{-3pt} & \textsc{pgd} & \textsc{fool} \\
\hline
\textsc{clean}  							& 84.0 & 83.5 & 82.7 & 32.4 & 17.0 & 0.012 \\
\textsc{wdecay} $5e\!-\!4$ 				& 84.3 & 84.1 & 83.4 & 26.5 & 11.0 & 0.015 \\
\textsc{pgd} 							& 81.8 & 81.9 & 81.8 & 62.7 & 57.8 & 0.050 \\
\textsc{fgsm} 							& 82.1 & 82.1 & 82.2 & 61.6 & 54.0 & 0.046 \\
\textsc{gn} $\lambda\!=\!0.075$  			& 80.3 & 80.2 & 80.0 & 52.2 & 46.7 & 0.040 \\
\textsc{sgr\_grad} $\lambda\!=\!0.1$  		& 80.2 & 80.0 & 80.2 & 53.7 & 48.1 & 0.042 \\
\textsc{sgr\_sign} $\lambda\!=\!0.5$	 		& 80.5 & 80.4 & 80.4 & 53.3 & 47.2 & 0.040 \\
\textsc{fgsm\_sgr} $\lambda\!=\!0.5$  		& 81.7 & 81.3 & 81.6 & 53.6 & 46.7 & 0.039 \\
\textsc{pgd\_sgr} $\lambda\!=\!0.5$ 			& 81.5 & 81.2 & 81.3 & 52.5 & 46.5 & 0.039 \\
\hline
\end{tabular} }
\end{tabular}
\caption{Test set and white-box attack accuracies. The different rows correspond to different models evaluated against the different white-box attacks indicated above each column (with first column showing test set accuracies). See Sec.~\ref{sec:whiteboxandtransferattacks} for a discussion of these numbers.}
\label{fig:cifarwhiteboxappendix}
\vspace{0.9cm}
{\small CIFAR10 $\epsilon=4$}\\[1mm]
{\tiny
\begin{tabular}{l c  c  c  c  c  c  c  c  c  c }
& & \multicolumn{9}{c}{ \textsc{train.\,meth.} } \\
\textsc{train.\,meth.} & \hspace{1mm}\textsc{test}\hspace{1mm} & \textsc{clean} & \textsc{wdecay} & \hspace{1mm}\textsc{pgd}\hspace{1mm} & \hspace{1mm}\textsc{fgsm}\hspace{1mm} & \hspace{2mm}\textsc{gn}\hspace{2mm} & \hspace{-1mm}\textsc{sgr\_grad}\hspace{-1mm} & \hspace{-1mm}\textsc{sgr\_sign}\hspace{-1mm} & \hspace{-1mm}\textsc{pgd\_sgr}\hspace{-1mm} & \hspace{-1mm}\textsc{fgsm\_sgr} \\
\hline
\textsc{clean}				 				& 84.3 & 13.9 & 59.0 & 75.2 & 74.6 & 62.6 & 62.7 & 61.7 & 64.4 & 61.9 \\
\textsc{wdecay}	 $5e\!-\!4$		 				& 85.2 & 68.7 & 9.2 & 75.6 & 75.4 & 61.6 & 61.4 & 59.6 & 63.2 & 61.5 \\
\textsc{pgd} 				 				& 81.9 & 80.9 & 80.4 & 56.1 & 76.4 & 77.1 & 77.7 & 77.1 & 77.3 & 77.3 \\
\textsc{fgsm}				 				& 82.2 & 81.2 & 80.3 & 75.2 & 51.3 & 76.8 & 77.5 & 76.8 & 77.0 & 77.0 \\
\textsc{gn}	 $\lambda\!=\!0.05$					& 83.3 & 80.0 & 78.0 & 74.3 & 75.2 & 37.0 & 69.7 & 68.9 & 70.2 & 69.4 \\
\textsc{sgr\_grad} $\lambda\!=\!0.03$			& 83.7 & 79.4 & 77.6 & 74.6 & 75.3 & 67.1 & 31.3 & 67.8 & 68.7 & 68.4 \\
\textsc{sgr\_sign} $\lambda\!=\!0.2$				& 84.2 & 80.0 & 77.8 & 75.1 & 75.8 & 68.3 & 69.4 & 33.7 & 69.3 & 68.1 \\
\textsc{pgd\_sgr} $\lambda\!=\!0.35$				& 83.2 & 79.4 & 77.7 & 74.2 & 74.9 & 68.6 & 69.6 & 68.3 & 36.2 & 68.9 \\
\textsc{fgsm\_sgr} $\lambda\!=\!0.25$			& 83.7 & 79.6 & 77.9 & 75.3 & 76.0 & 69.1 & 70.0 & 68.1 & 69.8 & 35.3 \\
\hline
\end{tabular} }
\\[5mm]
{\small CIFAR10 $\epsilon=8$}\\[1mm]
{\tiny
\begin{tabular}{l c  c  c  c  c  c  c  c  c  c }
& & \multicolumn{9}{c}{ \textsc{train.\,meth.} } \\
\textsc{train.\,meth.} & \hspace{1mm}\textsc{test}\hspace{1mm} & \textsc{clean} & \textsc{wdecay} & \hspace{1mm}\textsc{pgd}\hspace{1mm} & \hspace{1mm}\textsc{fgsm}\hspace{1mm} & \hspace{2mm}\textsc{gn}\hspace{2mm} & \hspace{-1mm}\textsc{sgr\_grad}\hspace{-1mm} & \hspace{-1mm}\textsc{sgr\_sign}\hspace{-1mm} & \hspace{-1mm}\textsc{pgd\_sgr}\hspace{-1mm} & \hspace{-1mm}\textsc{fgsm\_sgr} \\
\hline
\textsc{clean} 								& 84.3 & 11.2 & 27.7 & 61.6 & 68.1 & 27.8 & 27.9 & 26.5 & 30.3 & 27.9 \\
\textsc{wdecay} $5e\!-\!4$						& 85.2 & 44.5 & 8.0 & 63.7 & 68.6 & 25.6 & 26.2 & 24.2 & 27.7 & 26.5 \\
\textsc{pgd} 								& 77.7 & 76.2 & 75.0 & 40.5 & 71.3 & 66.6 & 67.9 & 67.3 & 67.1 & 66.9 \\
\textsc{fgsm}								& 81.2 & 75.9 & 74.0 & 60.3 & 13.2 & 62.9 & 65.5 & 63.8 & 63.8 & 63.8 \\
\textsc{gn} $\lambda\!=\!0.05$ 					& 83.3 & 75.0 & 70.1 & 63.9 & 70.3 & 11.2 & 45.9 & 44.0 & 47.0 & 45.6 \\
\textsc{sgr\_grad} $\lambda\!=\!0.03$ 			& 83.7 & 73.3 & 67.8 & 63.5 & 70.9 & 38.1 & 10.3 & 41.0 & 42.2 & 41.8 \\
\textsc{sgr\_sign} $\lambda\!=\!0.2$				& 84.2 & 74.3 & 67.8 & 65.3 & 70.9 & 40.5 & 43.6 & 10.0 & 42.8 & 40.5 \\
\textsc{pgd\_sgr} $\lambda\!=\!0.35$ 				& 83.2 & 74.9 & 69.5 & 65.3 & 70.5 & 43.6 & 46.6 & 43.7 & 10.9 & 44.1 \\
\textsc{fgsm\_sgr} $\lambda\!=\!0.25$ 			& 83.7 & 73.6 & 69.0 & 65.3 & 70.9 & 42.2 & 45.5 & 41.4 & 43.9 & 11.3 \\
\hline
\end{tabular} }
\\[5mm]
{\small MNIST $\epsilon=32$}\\[1mm]
{\tiny
\begin{tabular}{l c  c  c  c  c  c  c  c  c  c }
& & \multicolumn{9}{c}{ \textsc{train.\,meth.} } \\
\textsc{train.\,meth.} & \hspace{1mm}\textsc{test}\hspace{1mm} & \textsc{clean} & \textsc{wdecay} & \hspace{1mm}\textsc{pgd}\hspace{1mm} & \hspace{1mm}\textsc{fgsm}\hspace{1mm} & \hspace{2mm}\textsc{gn}\hspace{2mm} & \hspace{-1mm}\textsc{sgr\_grad}\hspace{-1mm} & \hspace{-1mm}\textsc{sgr\_sign}\hspace{-1mm} & \hspace{-1mm}\textsc{fgsm\_sgr}\hspace{-1mm} & \hspace{-1mm}\textsc{pgd\_sgr}  \\
\hline
\textsc{clean} 								& 99.4 & 24.7 & 94.2 & 98.4 & 98.7 & 95.8 & 95.4 & 95.7 & 97.1 & 96.2 \\
\textsc{wdecay} $2e\!-\!4$						& 99.4 & 97.7 & 48.3 & 98.1 & 98.7 & 95.4 & 95.2 & 96.4 & 97.4 & 97.0 \\
\textsc{pgd} 								& 99.5 & 99.3 & 99.3 & 98.2 & 99.2 & 99.1 & 99.1 & 99.3 & 99.3 & 99.2 \\
\textsc{fgsm}								& 99.5 & 99.2 & 99.1 & 99.0 & 96.0 & 98.8 & 98.9 & 99.0 & 99.1 & 99.0 \\
\textsc{gn} $\lambda\!=\!5.0$					& 99.4 & 98.8 & 98.0 & 98.5 & 98.9 & 88.9 & 96.3 & 97.9 & 98.3 & 98.0 \\
\textsc{sgr\_grad} $\lambda\!=\!0.1$	 			& 99.5 & 98.9 & 98.2 & 98.9 & 99.1 & 96.4 & 86.4 & 97.7 & 98.5 & 98.0 \\
\textsc{sgr\_sign} $\lambda\!=\!1.0$				& 99.5 & 98.8 & 98.3 & 98.8 & 98.9 & 97.5 & 97.2 & 86.9 & 98.1 & 97.9 \\
\textsc{fgsm\_sgr} $\lambda\!=\!0.5$				& 99.5 & 98.7 & 97.8 & 98.7 & 98.7 & 97.3 & 97.0 & 97.1 & 72.9 & 96.9 \\
\textsc{pgd\_sgr} $\lambda\!=\!1.0$				& 99.6 & 98.7 & 98.0 & 98.8 & 98.8 & 97.3 & 96.8 & 97.3 & 97.5 & 79.5 \\
\hline
\end{tabular} }
\\[5mm]
{\small MNIST $\epsilon=64$}\\[1mm]
{\tiny
\begin{tabular}{l c  c  c  c  c  c  c  c  c  c }
& & \multicolumn{9}{c}{ \textsc{train.\,meth.} } \\
\textsc{train.\,meth.} & \hspace{1mm}\textsc{test}\hspace{1mm} & \textsc{clean} & \textsc{wdecay} & \hspace{1mm}\textsc{pgd}\hspace{1mm} & \hspace{1mm}\textsc{fgsm}\hspace{1mm} & \hspace{2mm}\textsc{gn}\hspace{2mm} & \hspace{-1mm}\textsc{sgr\_grad}\hspace{-1mm} & \hspace{-1mm}\textsc{sgr\_sign}\hspace{-1mm} & \hspace{-1mm}\textsc{fgsm\_sgr}\hspace{-1mm} & \hspace{-1mm}\textsc{pgd\_sgr} \\
\hline
\textsc{clean} 								& 99.4 & 0.6 & 30.3 & 79.6 & 79.1 & 31.5 & 32.6 & 38.1 & 72.4 & 65.8 \\
\textsc{wdecay} $2e\!-\!4$ 					& 99.4 & 80.6 & 0.4 & 84.0 & 88.4 & 26.4 & 27.4 & 46.7 & 81.3 & 75.6 \\
\textsc{pgd} 								& 99.5 & 98.8 & 97.7 & 62.6 & 90.7 & 89.7 & 89.7 & 92.4 & 97.7 & 97.2 \\
\textsc{fgsm} 								& 99.5 & 98.0 & 93.7 & 76.8 & 38.6 & 81.8 & 81.8 & 84.0 & 94.6 & 92.1 \\
\textsc{gn} $\lambda\!=\!5.0$ 					& 99.4 & 96.3 & 86.5 & 91.3 & 96.4 & 9.3 & 55.0 & 79.5 & 93.2 & 90.7 \\
\textsc{sgr\_grad} $\lambda\!=\!0.1$ 				& 99.5 & 96.1 & 84.7 & 90.2 & 95.3 & 50.1 & 4.5 & 68.8 & 89.6 & 82.9 \\
\textsc{sgr\_sign} $\lambda\!=\!1.0$ 				& 99.5 & 95.1 & 83.6 & 88.8 & 93.4 & 59.0 & 52.7 & 2.4 & 85.3 & 82.7 \\
\textsc{fgsm\_sgr} $\lambda\!=\!0.5$ 				& 99.5 & 90.8 & 74.9 & 85.6 & 85.7 & 54.0 & 48.5 & 48.4 & 0.4 & 49.1 \\
\textsc{pgd\_sgr} $\lambda\!=\!1.0$ 				& 99.6 & 92.9 & 78.1 & 87.8 & 89.8 & 50.9 & 44.0 & 56.1 & 62.1 & 0.5 \\
\hline
\end{tabular} }
\caption{PGD transfer attack accuracies. The different rows correspond to different models evaluated on PGD adversarial perturbations generated from the corresponding models indicated above each column (with first column showing test set accuracies). Note that the PGD \& FGSM models were trained separately for each $\epsilon$. See Sec.~\ref{sec:whiteboxandtransferattacks} for a discussion of these numbers.}
\label{fig:transferappendix}
\end{center}
\end{table*}

\newpage

\begin{figure}[t]
  \begin{center}
    \begin{tabular}{@{}c@{}c@{}}
    \adjincludegraphics[width=0.3\columnwidth, trim={{0.1\width} {0.29\height} {0.5\width} {0.11\height}},clip]{/AdversarialPerturbations/{2018_05_16_1412_mnist_clean_32.0eps_grid_pgd}.png} &
    \,\adjincludegraphics[width=0.293\columnwidth, trim={{0.51\width} {0.29\height} {0.1\width} {0.11\height}},clip]{/AdversarialPerturbations/{2018_05_16_1413_mnist_clean_64.0eps_grid_pgd}.png} 
    \end{tabular}
    \caption{PGD samples for clean model on MNIST. (Left half) for $\epsilon=32$, (right half) for $\epsilon=64$.
    Top row shows original test set images sorted according to their true class label, middle shows adversarially perturbed samples, while the bottom row shows
the adversarial perturbations rescaled to fit the full range of rgb values (their actual range and hence saliency is smaller).
}
    \label{fig:PGDmnistamples}
\vspace{1cm}
    \begin{tabular}{@{}c@{}c@{}}
    \adjincludegraphics[width=0.3\columnwidth, trim={{0.1\width} {0.29\height} {0.5\width} {0.11\height}},clip]{/AdversarialPerturbations/{2018_05_15_1554_cifar10_clean_4.0eps_grid_pgd}.png} \hspace{-1mm}&\hspace{-1mm}
    \adjincludegraphics[width=0.293\columnwidth, trim={{0.51\width} {0.29\height} {0.1\width} {0.11\height}},clip]{/AdversarialPerturbations/{2018_05_15_1555_cifar10_clean_8.0eps_grid_pgd}.png} 
    \end{tabular}
        \caption{PGD samples for clean model on CIFAR10. (Left half) for $\epsilon=4$, (right half) for $\epsilon=8$.
    Top row shows original test set images sorted according to their true class label, middle shows adversarially perturbed samples, while the bottom row shows
the adversarial perturbations rescaled to fit the full range of rgb values (their actual range and hence saliency is smaller).
}
    \label{fig:PGDcifarsamples}
  \end{center}
\end{figure}

\subsection{Classifier Architectures}
\label{sec:ClassifierArchitecturesAppendix}

All networks were trained for $50$ epochs using the Adam optimizer \cite{kingma2014adam} with learning rate $0.001$ and minibatch size $128$. 

\begin {table}[h!]
\begin{center}
\begin{tabular}{| c | c |}
  \hline
   \multicolumn{2}{|c|}{\bf Classifier} \\
  \hline \hline
   \multirow{2}{*}{Feature Block} &  Conv2D (filter size: $3\times3$, feature maps: params[0], stride: $1\times1$)  \\ 
    & ReLU  \\ 
   \hline
   
   \multirow{2}{*}{Feature Block} &  Conv2D (filter size: $3\times3$, feature maps: params[1], stride: $1\times1$)  \\ 
    & ReLU  \\ 
   \hline
   
   {MaxPooling} &  MaxPool (pool size: (2,2))  \\ 
   \hline  
   
   \multirow{2}{*}{Feature Block} &  Conv2D (filter size: $3\times3$, feature maps: params[2], stride: $1\times1$)  \\ 
    & ReLU  \\ 
   \hline
   
   \multirow{2}{*}{Feature Block} &  Conv2D (filter size: $3\times3$, feature maps: params[3], stride: $1\times1$)  \\ 
    & ReLU  \\ 
   \hline
   
   {MaxPooling} &  MaxPool (pool size: (2,2))  \\ 
   \hline  
   
   {Fully-Connected} &  Dense (units: params[4])  \\ 
   \hline  
   
   {Fully-Connected} &  Dense (units: params[5])  \\ 
   \hline  
   
   \multirow{2}{*}{Fully-Connected} &  Dense (units: params[6]) \\
   & Softmax  \\
   \hline  
   
\end{tabular}
\vspace{5mm}
\caption{Classifier Architectures: CIFAR10 params=$[64, 64, 128, 128, 256, 256, 10]$, MNIST params$=[32, 32, 64, 64, 200, 200, 10]$.
Our models are identical to those used in \cite{carlini2017towards, papernot2016distillation}.}
\end{center}
\end{table}

\end{document}